\documentclass[11pt]{article}

\usepackage[preprint]{acl}

\usepackage{times}
\usepackage{latexsym}
\usepackage{amsmath}
\usepackage{amssymb}
\usepackage{booktabs}
\usepackage{multirow}
\usepackage{amsthm}
\usepackage[T1]{fontenc}

\usepackage[utf8]{inputenc}

\usepackage{microtype}

\usepackage{inconsolata}

\usepackage{graphicx}

\title{Diagnosing Harmful Continuation in
 Answer-Correct \\ Long-CoT Training Traces}

\author{
Chen He$^{1*}$ \quad
Yuhao Wu$^{2*}$ \quad
Lei Wang$^{3}$ \quad
Wenxuan Zhang$^{2}$ \quad
Fumin Shen$^{1\dagger}$ \\
$^{1}$University of Electronic Science and Technology of China \\
$^{2}$Singapore University of Technology and Design \quad
$^{3}$Singapore Management University \\
$^{*}$Equal contribution  $\quad \dagger $ Corresponding author \\
}

\begin{document}
\maketitle
\begin{abstract}
Long chain-of-thought (CoT) traces are widely used as supervision for reasoning-oriented LLM SFT, yet answer-correct traces can still lead to markedly different fine-tuning outcomes.
We study post-conclusion continuation in answer-correct long-CoT data: a continuation where the answer appears sufficiently supported, but the trace continues with additional reasoning that remains in the supervised target.
To test its training effect, we use a delete-only editor to construct answer-preserving suffix removal and compare CoT-based SFT on the original and processed traces.
We observe improved SFT outcomes after removing the editor-identified post-conclusion continuation, suggesting that this continuation is harmful to training in our setting.
We therefore refer to this empirically supported phenomenon as \textit{harmful continuation}.
Beyond this intervention, we further characterize the removed post-conclusion continuation through uncertainty and hidden-state progress.
We observe persistent local uncertainty together with weakened terminal-directional progress, forming an uncertainty--geometry mismatch.
Finally, we instantiate \textit{\textbf{H}armful \textbf{C}ontinuation \textbf{C}ut (HCC)}, a lightweight boundary proxy that approximates the editor-identified post-conclusion continuation boundary.
\end{abstract}

\section{Introduction}
Long chain-of-thought traces have become important training targets for reasoning models \citep{wei2022chain,luo2025deconstructing,ou2025empowering}.
They are used not only in supervised fine-tuning \cite{ou2025empowering}, but also in reasoning-oriented continued training and as cold-start data before reinforcement learning \cite{wang2026beyond}.
Unlike final-answer annotations, long-CoT traces expose full reasoning trajectories that models are encouraged to imitate during training.
This makes CoT trace quality a central issue for reasoning training where the training target specifies not only what answer to produce, but also what trajectory should be treated as learnable reasoning behavior.

Prior work has already shown that answer-correct CoT traces can differ substantially in training utility.
The style and structure of source trajectories can affect the generalization of SFT models \citep{tian2025not,zhang2025best,li2025small}, and recent studies further connect trace compatibility, reasoning patterns, and learnability to downstream outcomes \citep{liu2026long,li2026role}.
However, most existing methods remain confined to trace selection, prefix selection, or externally guided rewriting, which leaves the internal failure mode of answer-correct traces under-explained.
As a result, they do not characterize where useful reasoning may end and begin, or why such a phase can be associated with weaker SFT despite preserving answer correctness.

To address this gap, we take a diagnostic view of answer-correct long-CoT traces.
We seek a trace-internal diagnostic explanation for why answer-correct traces may differ in training utility, rather than assuming that a long reasoning trace is uniformly useful once its final answer is correct.
Our goal is not to claim that every long tail is harmful, nor to treat length as the central issue.
Instead, we ask whether some traces enter a low-value post-conclusion continuation: the answer is already sufficiently supported, but subsequent reasoning remains locally costly while showing weak hidden-state progress.
From the uncertainty perspective, we observe that some post-conclusion continuation remains locally costly or unstable, suggesting that the trace continues to explore after evaluator-based answer support has largely saturated.
From the geometric perspective, this continued exploration shows weakened terminal-directional hidden-state progress.
We refer to this hypothesized low-value phase as post-conclusion continuation before evaluating its downstream training effect.
When answer-preserving removal of this continuation improves SFT outcomes, we call the empirically supported training-unfavorable case harmful continuation.

To examine the training relevance of this post-conclusion continuation, we use a delete-only editor as an operational intervention tool.
The editor does not rewrite the trace; it only removes post-conclusion suffixes while preserving the original prefix and final answer.
This allows us to test whether answer-preserving removal of the post-conclusion continuation improves SFT outcomes.
Motivated by this diagnosis, we instantiate \textit{\textbf{H}armful \textbf{C}ontinuation \textbf{C}ut (HCC)}, a lightweight boundary proxy.
HCC uses a frozen Qwen2.5-0.5B-Instruct backbone with a cut head to extract sentence-level reasoning states and approximate the editor-identified post-conclusion continuation.

Our contributions can be summarized as follows:
\begin{itemize}
\item We formulate post-conclusion continuation in answer-correct long-CoT traces without assuming inherent harmfulness.
\item We show through answer-preserving suffix removal that the removed continuation is training-unfavorable in our SFT settings.
\item We characterize the removed continuation with an uncertainty--geometry mismatch and propose HCC as a proxy to remove it.
\end{itemize}

\section{Related Work}
\noindent\textbf{Long-CoT reasoning training.}
Long-CoT traces have become important for post-training pipelines.
While early pipelines often treat answer-correct traces as directly usable supervision, recent studies show that correctness alone does not determine their training value.
Work on data selection, trace compatibility, and informative alignment \cite{yang2026reasoning,zhang2025best,chandra2025shape} suggests that only a subset of answer-correct traces provides beneficial supervision.
Other approaches modify or shorten reasoning traces through sequence truncation, prefix optimization, adaptive prefix alignment, robustness to partial reasoning, or length-aware training \cite{chen2025distilling,sun2026well,liu2026long,silvestri2026learning,xu2025chain,luo2025o1,ma2025cot}.
However, these methods mainly operate in a heuristic manner.
They do not directly characterize where useful reasoning may end, where useless reasoning may begin, or why this reasoning can be associated with weaker SFT supervision despite answer correctness.

\noindent\textbf{Properties of long-CoT trajectories.}
Recent work increasingly treats long-CoT traces as structured reasoning trajectories rather than flat text sequences.
Studies on overthinking show that long-reasoning models may repeatedly verify intermediate conclusions or continue reasoning without meaningful gain \cite{chen2025not}.
Complementary analyses examine global reasoning patterns, trajectory geometry, and step-level anchors that contribute to actual progress \cite{jiang2025makes,ballon2026probing,bogdan2025thought,yang2025demystifying}.
Most closely related to our motivation, \citet{li2026role} connects trajectory properties to downstream SFT outcomes and shows that different reasoning patterns can lead to different generalization behavior.
Our work builds on this trajectory-level view, but focuses on a post-conclusion continuation inside answer-correct traces: continuation that remains uncertain or costly while showing weakened terminal-directional progress.

\section{Operational Partition and Diagnostics of Post-Conclusion Continuation}
\label{sec:illustration}

\subsection{Data Construction}
\label{subsec:data_construction}
In this section, we do not assume that editor-removed sentences are ground-truth harmful continuation.
Instead, we use the delete-only editor to construct an operational partition for diagnosing post-conclusion suffixes.
The resulting groups are used to reveal statistical signatures of a possible low-value phase, rather than to define harmfulness by editor decisions alone.
We use Qwen3-235B-A22B-Instruct-2507 \cite{qwen3technicalreport} and DeepSeek-R1-V3.2 \cite{guo2025deepseek} to generate trajectories, and sample 4,780 answer-correct long-CoT solution trajectories from the OpenR1-Math-220k dataset.
These trajectories serve as the original CoT training traces.
For simplicity, we use $\{T\}_{Q}$ and $\{T\}_{R}$ to refer to the two sets of trajectories from the two models, respectively.
We then employ Qwen3.5-27B \cite{qwen3technicalreport} as a delete-only offline editor to expose post-conclusion continuation for empirical analysis.
Given a trajectory from $\{T\}_{Q}$ or $\{T\}_{R}$, the editor marks the post-conclusion sentences that can be removed while preserving the reasoning necessary to recover the final answer.

\begin{figure}[htb!]
  \centering
  \includegraphics[width=0.48\textwidth]{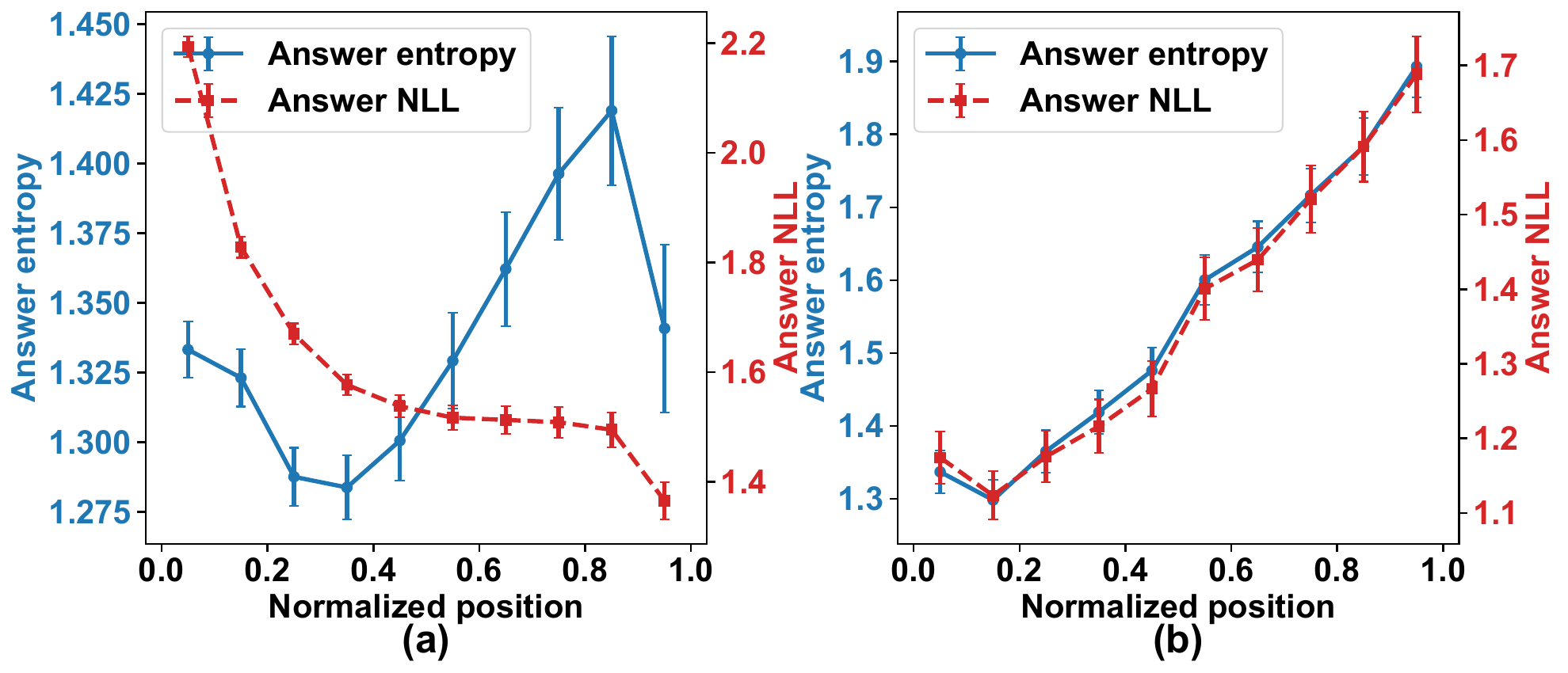}
  \vspace{-4mm}
\caption{Evaluator-based uncertainty diagnostics as reasoning segments are progressively added for (a) retained reasoning and (b) editor-removed continuation.}
  \vspace{-5mm}
  \label{fig:process-uncer}
\end{figure}

\paragraph{Operational groups.}
We divide each edited trajectory into two operational groups.
The first group is the \emph{retained reasoning}, namely the editor-preserved portion that supports the final answer.
The second group is the \emph{editor-removed continuation}, namely the post-conclusion continuation marked as removable by the offline editor.
At this stage, these terms refer to operational groups rather than predefined theoretical labels.

\subsection{Uncertainty View}
\label{subsec:uncertainty_view}
In this section, we aim to answer the following question:
\textit{Does the post-conclusion continuation continue to improve evaluator-based final-answer recoverability, or does answer support appear to saturate while local uncertainty remains high?}

\noindent \textbf{Comparison protocols.}
We analyze uncertainty at both answer and sentence levels.
At the answer level, we progressively append reasoning sentences along the same complete response trajectory and compute prefix-conditioned final-answer entropy and NLL.
These quantities should be interpreted as evaluator-based diagnostics of answer recoverability, rather than direct measurements of causal reasoning contribution.
For segment-wise visualization, positions are normalized separately within retained reasoning and the subsequent editor-removed continuation.
At the sentence level, we use sentence entropy and sentence NLL to measure local predictive difficulty.
For boundary-level analysis, we track $K_1$, $K_T$, $C_1$, and $C_T$, denoting the first and last sentences of retained reasoning and editor-removed continuation, and compare both local uncertainty changes and answer-NLL reduction. 
The detailed protocols are described in the Appendix.

\begin{figure}[htb!]
  \centering
  \begin{minipage}{0.98\linewidth}
    \centering
    \includegraphics[width=\linewidth]{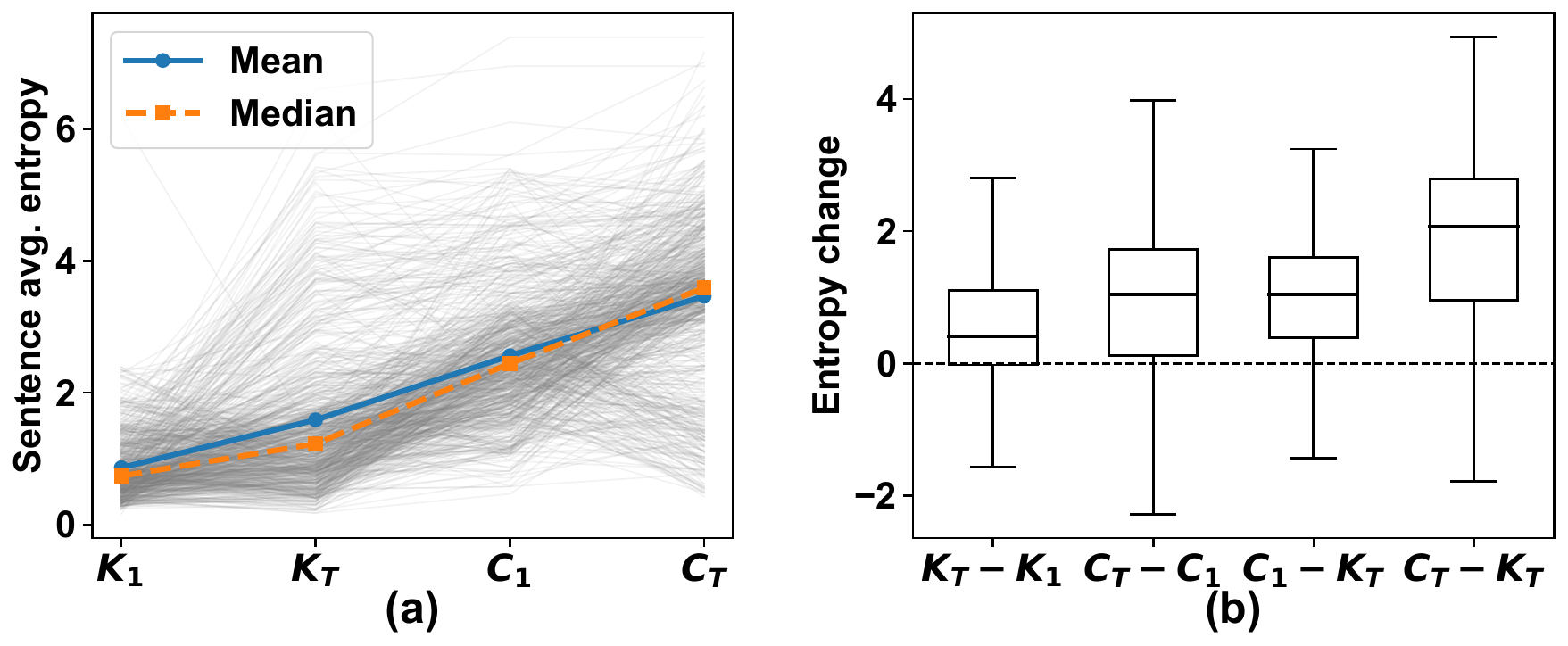}
  \end{minipage}
  \vspace{0.5em}
  \begin{minipage}{0.98\linewidth}
    \centering
    \includegraphics[width=\linewidth]{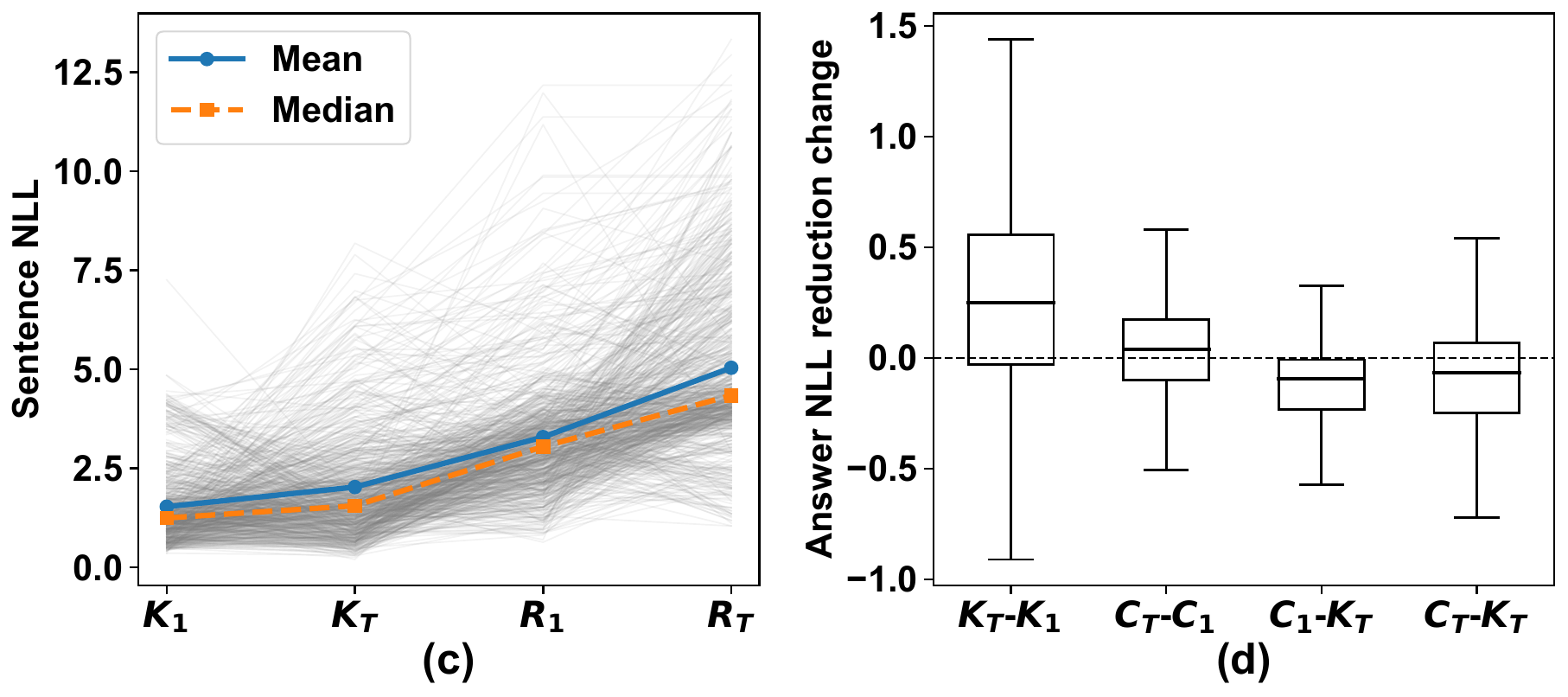}
  \end{minipage}
  \vspace{-5mm}
\caption{
Boundary-level diagnostic changes around the editor-identified post-conclusion continuation: (a) sentence entropy, (b) entropy change, (c) sentence NLL, and (d) answer-NLL reduction change.
}
  \vspace{-2mm}
  \label{fig:uncertainty-view}
\end{figure}

\begin{figure}[htb!]
  \centering
  \includegraphics[width=0.48\textwidth]{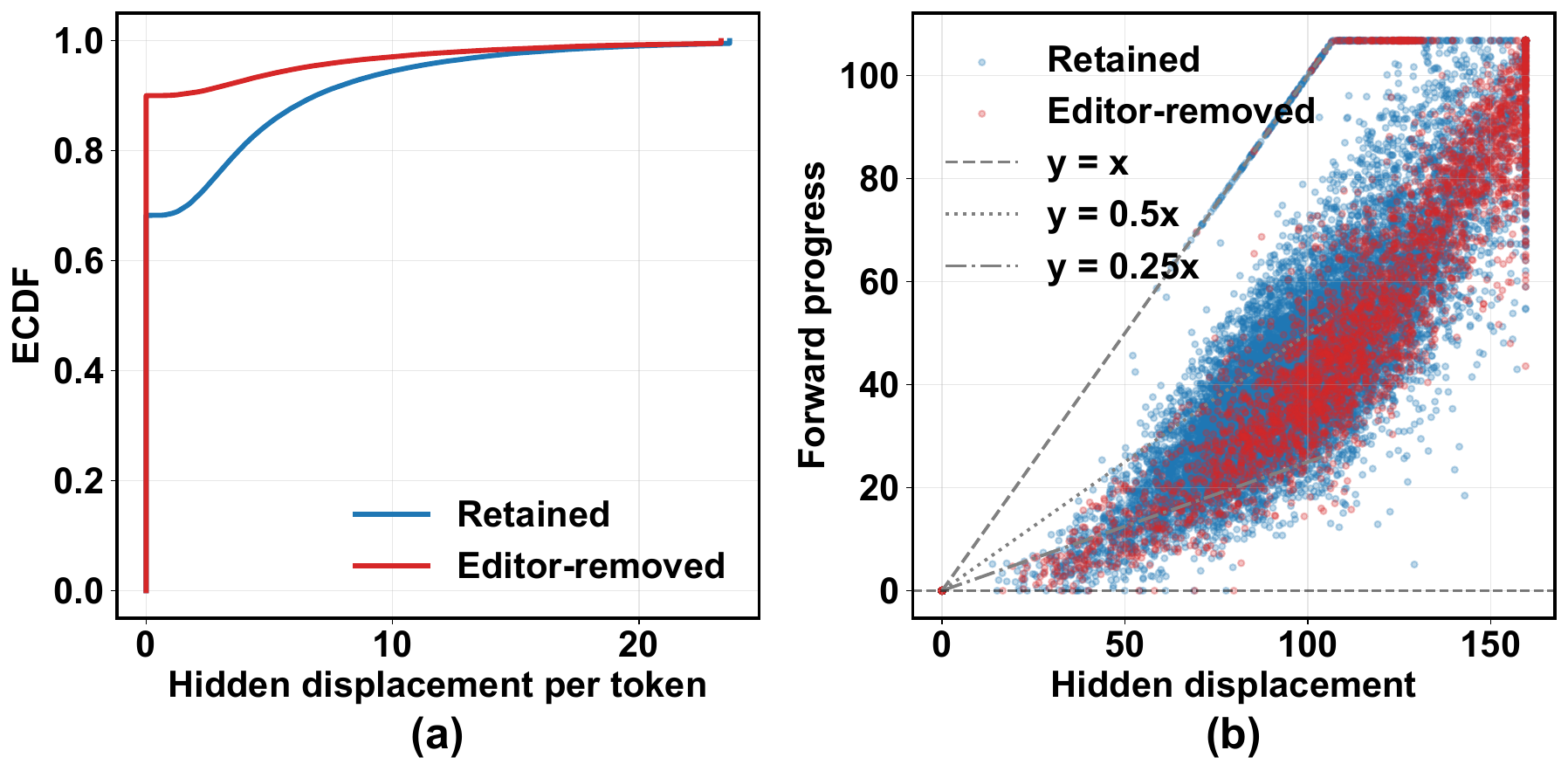}
  \vspace{-6mm}
\caption{
Operational hidden-state progress of retained reasoning and editor-removed continuation:
(a) ECDF of token-normalized hidden displacement and
(b) hidden displacement versus forward progress.
}
  \label{fig:geometry-r1}
  \vspace{-5mm}
\end{figure}

\noindent \textbf{Answer uncertainty dynamics.}
Figure~\ref{fig:process-uncer} shows how answer-level uncertainty changes as reasoning sentences are progressively appended to the same complete response.
For visualization, the x-axis is normalized within its corresponding segment, with Figure~\ref{fig:process-uncer} (a) showing retained reasoning and Figure~\ref{fig:process-uncer} (b) showing the subsequent editor-removed continuation.
In retained reasoning, answer entropy changes non-monotonically but does not exhibit a persistent increase, while answer NLL steadily decreases as more useful reasoning is added.
This suggests that the retained segment improves evaluator-based final-answer recoverability even when intermediate reasoning involves local exploration or verification.
In contrast, once the trace enters editor-removed continuation, both answer entropy and answer NLL increase as more post-conclusion content is appended.
This suggests that the continuation does not consistently improve evaluator-based answer recoverability, but instead introduces a higher-uncertainty state after the answer has been sufficiently supported.

\begin{table*}[htb!]
\centering
\small
\setlength{\tabcolsep}{3pt}
\renewcommand{\arraystretch}{1.1}
\caption{
Paired per-sample comparison of operational hidden-state progress between editor-removed continuation and retained reasoning.
The paired difference is defined as $\Delta=\text{Removed mean}-\text{Retained mean}$.
}
\vspace{-2mm}
\label{tab:paired_geometry_summary}
\begin{tabular}{cccccc}
\toprule
\textbf{Metric} &
\textbf{Removed mean} &
\textbf{Retained mean} &
\textbf{Removed lower} &
\textbf{Removed higher} &
\textbf{95\% CI of $\Delta$} \\
\midrule
Hidden displacement              & 21.91 & 44.92 & 0.79 & 0.16 & $[-23.85,\,-22.19]$ \\
Forward progress                 & 10.79 & 20.50 & 0.79 & 0.17 & $[-10.15,\,-9.31]$ \\
Progress efficiency              & 0.09  & 0.19  & 0.79 & 0.17 & $[-0.11,\,-0.10]$ \\
Hidden displacement / token      & 1.65  & 2.83  & 0.79 & 0.16 & $[-1.24,\,-1.11]$ \\
Forward progress / token         & 0.84  & 1.28  & 0.79 & 0.17 & $[-0.47,\,-0.40]$ \\
Curvature                        & 1.09  & 1.17  & 0.73 & 0.27 & $[-0.09,\,-0.08]$ \\
\bottomrule
\end{tabular}
\vspace{-4mm}
\end{table*}

\noindent \textbf{Boundary-level mismatch.}
Figure~\ref{fig:uncertainty-view} examines local uncertainty and evaluator-based answer-support changes on the boundary between retained reasoning and editor-removed continuation.
From $K_1$ to $K_T$, sentence entropy and sentence NLL increase, but the answer-NLL reduction also becomes stronger, suggesting that local uncertainty within retained reasoning can still accompany improved answer recoverability under the evaluator.
The transition from $K_T$ to $C_1$ shows a different pattern.
Local uncertainty rises at the beginning of editor-removed continuation, while the gain in answer support no longer increases correspondingly.
From $C_1$ to $C_T$, this high-uncertainty regime is maintained or amplified, but the continuation does not provide stable additional answer-NLL reduction.
The candidate low-value pattern therefore emerges when increased local prediction difficulty is no longer matched by consistent improvements in evaluator-based final-answer recoverability.

\subsection{Geometric View}
\label{subsec:geometric_view}

Following the uncertainty analysis, another question arises: \textit{Does the increased predictive uncertainty of post-conclusion continuation translate into effective hidden-state progress?}

\noindent \textbf{Comparison protocols.}
Following prior work on the geometry of Transformer hidden representations \cite{valeriani2023geometry,gurnee2024language} and trajectory-level analyses of long-CoT reasoning \cite{jiang2025makes}, we use sentence-boundary hidden states as an operational proxy for reasoning-state evolution.
Specifically, hidden displacement measures the magnitude of representation change between consecutive reasoning steps, while forward progress measures the component of this change aligned with the terminal direction of the analyzed trace.
These metrics characterize representation-level state movement and terminal-directional progress under an operational proxy.
Because the terminal direction is derived from the observed trace representation, forward progress should be interpreted as an operational terminal-directional proxy rather than a ground-truth answer direction.
We further compute progress efficiency, defined as the ratio between forward progress and hidden displacement, to measure how effectively local state movement is converted into directional progress.
To control for sentence length, we also report token-normalized variants of hidden displacement and forward progress.
Curvature is included as an auxiliary diagnostic of directional change, while displacement, forward progress, and efficiency serve as the primary geometric indicators.
The detailed protocols are described in Appendix.

\noindent \textbf{Distributional geometric tendency.}
Figure~\ref{fig:geometry-r1}(a) shows that many sentences in both groups have near-zero token-normalized hidden displacement, suggesting that fine-grained geometric scores should not be used as hard sentence-level deletion criteria.
Nevertheless, retained reasoning is shifted toward larger displacement, indicating stronger state movement per token than editor-removed continuation.
Figure~\ref{fig:geometry-r1}(b) shows a similar pattern.
Although the two groups overlap substantially in the scatter space, retained reasoning more often exhibits larger hidden displacement and stronger forward progress.
Thus, geometry provides a distributional signal of useful reasoning progress rather than a pointwise separation rule.

\noindent \textbf{Paired evidence.}
Table~\ref{tab:paired_geometry_summary} gives a paired per-sample comparison and shows that editor-removed continuation has weaker operational hidden-state progress than retained reasoning.
For hidden displacement, the removed mean is much lower than the retained mean ($21.91$ vs. $44.92$), and the paired difference is consistently negative, with $79\%$ of samples showing lower values for the removed segment.
Forward progress shows the same trend, suggesting that the removed continuation advances less under the terminal-directional proxy.
These gaps remain after token normalization, where both hidden displacement per token and forward progress per token are lower for editor-removed continuation.
By contrast, curvature has only a small absolute gap, suggesting that the post-conclusion continuation is associated mainly with weaker displacement and weaker terminal-directional progress, rather than with a curvature pattern.

\subsection{Uncertainty--Geometry Mismatch}
\label{subsec:summary}
The above analyses reveal a consistent diagnostic pattern in the editor-identified post-conclusion continuation.
From the uncertainty view, this suffix often remains locally costly or unstable, while evaluator-based answer recoverability no longer improves consistently.
From the geometric view, the same suffix shows weaker hidden displacement and weaker terminal-directional progress than retained reasoning.
We refer to this pattern as an uncertainty--geometry mismatch. 
Importantly, this mismatch is not used to define harmfulness and does not by itself prove causal training harm. 
Instead, it characterizes the editor-removable post-conclusion continuation that is later tested through downstream SFT intervention.

\section{Method}
\label{sec:method}

Our diagnostic analysis suggests that editor-removed post-conclusion continuation is associated with persistent local uncertainty and weak terminal-directional hidden-state progress.
Based on this operational pattern, we instantiate a lightweight boundary proxy termed Harmful Continuation Cut.

\subsection{Boundary Proxy}
\label{subsec:formulation}

Given a question $q$ and a verified source trajectory, we separate the sentence-level reasoning trace from the final answer.
Let $r = (r_1, r_2, \dots, r_T)$ denote the reasoning trace, and let $a^*$ denote the final answer.
Let $c^\ast \in \{0,\dots,T\}$ denote the editor-identified post-conclusion continuation boundary used for supervision.
Our goal is to learn a lightweight proxy that predicts this boundary, so that the retained prefix $r_{\le c^\ast}$ is kept while the post-conclusion continuation $r_{>c^\ast}$ is removed.

We encode the question and reasoning trace with a frozen causal language model and extract a hidden representation at each sentence boundary, obtaining $h_t \in \mathbb{R}^{D}$.
This representation denotes the model state after consuming the prefix up to sentence $r_t$.
We then pass the sentence-level states through a shared sequence encoder:
\begin{equation}
  \small
  \tilde{h_t} = \mathrm{SeqEnc}(h_{1:T})_t .
\end{equation}
The contextualized representation $\tilde{h_t}$ is used as the common input for latent regularization and uncertainty--geometry diagnostic estimation.

\subsection{Sequential Latent Regularization}
\label{subsec:ib}

HCC uses a sequential variational latent representation to regularize sentence-level boundary states.
For each contextualized sentence state $\tilde{h_t}$, we define a posterior latent distribution:
\begin{equation}
  \small
  q_\phi(z_t \mid \tilde{h_t})=\mathcal{N}(\mu_t,\Sigma_t),
\end{equation}
where the mean and variance are predicted from $\tilde{h_t}$.
We also define a sequential prior from the previous contextual state:
\begin{equation}
  \small
  p_\eta(z_t \mid \tilde{h_{t-1}})
  =
  \mathcal{N}(\mu^p_t,\Sigma^p_t).
\end{equation}
The sampled latent variable is projected back to the boundary-prediction space:
\begin{equation}
  \small
  b_t = f_{\mathrm{lat}}(z_t).
\end{equation}
The latent representation is regularized by:
\begin{equation}
  \small
  \mathcal{L}_{\mathrm{KL}}
  =
  \sum_{t=1}^{T}
  D_{\mathrm{KL}}
  \left(
  q_\phi(z_t \mid \tilde{h_t})
  \,\|\, 
  p_\eta(z_t \mid \tilde{h_{t-1}})
  \right).
\end{equation}
This term provides compact latent regularization for boundary prediction, rather than a hard information bottleneck or an explicit answer-prediction objective.

\subsection{Uncertainty--Geometry Diagnostic Estimation}
\label{subsec:mismatch}

Post-conclusion continuation may remain locally costly or unstable while contributing limited hidden-state progress.
To capture this uncertainty--geometry mismatch, HCC jointly estimates a local uncertainty signal and an operational progress signal from $\tilde{h_t}$.

For uncertainty, we define a scalar target $T_t$ from source-trace sentence-level statistics such as entropy, NLL, or log-perplexity.
HCC estimates this signal as:
\begin{equation}
  \small
  s^{\mathrm{ent}}_t,\hat T_t = f_{\mathrm{ent}}(\tilde{h_t}),
\end{equation}
where $s^{\mathrm{ent}}_t$ is the uncertainty-aware context vector used for boundary prediction, and $\hat T_t$ is the scalar uncertainty estimate.
The regression loss is:
\begin{equation}
  \small
  \mathcal{L}_{\mathrm{ent}}
  =
  \sum_{t=1}^{T}
  \mathrm{Huber}(\hat T_t, T_t).
\end{equation}

For geometry, we define a scalar progress target $G_t$ from hidden-state movement statistics.
HCC estimates this signal as:
\begin{equation}
  \small
  s^{\mathrm{geo}}_t,\hat G_t = f_{\mathrm{geo}}(\tilde{h_t}),
\end{equation}
where $s^{\mathrm{geo}}_t$ is the progress-aware context vector used for boundary prediction, and $\hat G_t$ is the scalar progress estimate.
The regression loss is:
\begin{equation}
  \small
  \mathcal{L}_{\mathrm{geo}}
  =
  \sum_{t=1}^{T}
  \mathrm{Huber}(\hat G_t, G_t).
\end{equation}

Together, these estimates provide a unified diagnostic representation of whether a post-conclusion sentence remains locally uncertain while showing weak operational progress.

\subsection{Mismatch-Aware Boundary Prediction}
\label{subsec:objective}

HCC fuses the latent regularization signal with the uncertainty--geometry diagnostic representation in a shared boundary-prediction space:
\begin{equation}
  \small
  m_t =
  \mathrm{LN}
  \left(
  b_t
  +
  \alpha_{\mathrm{geo}} s^{\mathrm{geo}}_t
  +
  \alpha_{\mathrm{ent}} s^{\mathrm{ent}}_t
  \right),
\end{equation}
where $\alpha_{\mathrm{geo}}$ and $\alpha_{\mathrm{ent}}$ are learnable scalar gates.
Thus, HCC does not concatenate the raw contextual state directly into the final cut representation.
Instead, the scalar estimates $\hat T_t$ and $\hat G_t$ are trained with auxiliary losses, while their context vectors contribute to boundary prediction.

For cut prediction, we prepend a learned beginning-of-sequence state $m_0$ and compute:
\begin{equation}
  \small
  \pi_t = \mathrm{CutHead}(m_t),
  \quad t=0,\dots,T.
\end{equation}
Here, $\pi_t$ denotes the logit of choosing sentence $t$ as the last retained sentence.
Given the editor-identified boundary $c^*$, the cut head is trained with:
\begin{equation}
  \small
  \mathcal{L}_{\mathrm{cut}}
  =
  - \log
  \frac{\exp(\pi_{c^*})}
  {\sum_{j=0}^{T}\exp(\pi_j)}.
\end{equation}

We also train a sentence-level deletion head to match the deletion labels produced by the offline editor.
Let $y_t \in \{0,1\}$ denote whether sentence $r_t$ should be deleted.
The deletion probability is:
\begin{equation}
  \small
  \hat y_t = \mathrm{DelHead}(m_t),
\end{equation}
with the binary deletion loss:
\begin{equation}
  \small
  \mathcal{L}_{\mathrm{del}}
  =
  -\sum_{t=1}^{T}
  \left[
  y_t \log \hat y_t
  +
  (1-y_t)\log(1-\hat y_t)
  \right].
\end{equation}

The overall training objective is:
\begin{equation}
  \small
  \begin{aligned}
  \mathcal{L}
  =
  &\mathcal{L}_{\mathrm{cut}}
  + \lambda_{\mathrm{del}}\mathcal{L}_{\mathrm{del}}
  + \lambda_{\mathrm{KL}}\mathcal{L}_{\mathrm{KL}} \\
  &+ \lambda_{\mathrm{ent}}\mathcal{L}_{\mathrm{ent}}
  + \lambda_{\mathrm{geo}}\mathcal{L}_{\mathrm{geo}} .
  \end{aligned}
\end{equation}

\section{Experiments}
\subsection{Experimental Setup}
\noindent \textbf{Datasets.}
We use the same trajectories as in Section~\ref{sec:illustration}, also denoted as $\{T\}_{Q}$ and $\{T\}_{R}$.
For HCC training, we use the 500M-scale \texttt{Qwen2.5-0.5B-Instruct} as a frozen backbone and train lightweight prediction heads to learn the editor-identified removable post-conclusion continuation boundary.
To test whether the learned deletion rule transfers across nearby source-model families rather than only memorizing a single source style, we construct the train--validation split across source models.
Specifically, we use $\{T\}_{Q}$ to train HCC, then process $\{T\}_{R}$ with the trained HCC and SFT the baseline model on the processed $\{T\}_{R}$, and vice versa.


\begin{table*}[htb!]
  \centering
  \small
  \setlength{\tabcolsep}{5pt}
  \renewcommand{\arraystretch}{1.05}
  \caption{Main results using different long-CoT trajectories. Len. denotes the average response length.}
  \label{tab:main_results}
  \begin{tabular}{ccccccccccc}
    \toprule
    \multirow{2}{*}{\textbf{Method}} 
    & \multicolumn{5}{c}{SFT on $\{T\}_{Q}$}
    & \multicolumn{5}{c}{SFT on $\{T\}_{R}$} \\
    \cmidrule(lr){2-6} \cmidrule(lr){7-11}
    & \textbf{MATH500} & \textbf{AMC23} & \textbf{GSM8K} & \textbf{Avg.} & \textbf{Len.}
    & \textbf{MATH500} & \textbf{AMC23} & \textbf{GSM8K} & \textbf{Avg.} & \textbf{Len.} \\
    \midrule

    \multicolumn{11}{c}{\texttt{Backbone: LLaMA3.2-3B-Instruct}} \\
    \midrule
    Vanilla
    & 29.8 & 10.0 & 69.0 & 36.3 & 3478.8
    & 38.4 & 15.0 & 69.9 & 41.1 & 4967.1 \\
    Heuristic
    & 34.0 & 15.0 & 71.5 & 40.2 & 3430.8
    & 40.0 & 15.0 & 70.1 & 41.7 & 5172.1 \\
    Editor
    & \underline{41.8} & \textbf{20.0} & \textbf{75.4} & \textbf{45.7} & 1906.1
    & \textbf{51.2} & \textbf{22.5} & \underline{75.9} & \textbf{49.8} & 1942.1 \\
    HCC
    & \textbf{43.2} & \underline{17.5} & \underline{75.1} & \underline{45.2} & 2010.1
    & \underline{47.6} & \textbf{22.5} & \textbf{77.8} & \underline{49.3} & 1934.4 \\

    \midrule
    \multicolumn{11}{c}{\texttt{Backbone: Qwen2.5-Math-7B-Instruct}} \\
    \midrule
    Vanilla
    & \textbf{85.8} & \underline{62.5} & 95.8 & \underline{81.4} & 519.6
    & 85.6 & 62.5 & 95.8 & 81.3 & 510.6 \\
    Heuristic
    & 82.8 & 57.5 & 95.8 & 78.7 & 526.1
    & 84.4 & 62.5 & 95.7 & 80.9 & 498.5 \\
    Editor
    & \underline{84.4} & \textbf{70.0} & \textbf{96.2} & \textbf{83.5} & 499.0
    & \underline{86.4} & \textbf{67.5} & \underline{96.0} & \textbf{83.3} & 443.2 \\
    HCC
    & 82.6 & \underline{62.5} & \underline{96.1} & 80.4 & 497.9
    & \textbf{86.6} & \underline{65.0} & \textbf{96.2} & \underline{82.6} & 455.9 \\

    \bottomrule
  \end{tabular}
\end{table*}

\noindent \textbf{Evaluation metrics.}
As for the comparison methods, we include 4 different methods to process the original trajectories, including (1) Vanilla, indicating that the original trajectories are directly used for SFT without any processing; (2) Editor, indicating that the trajectory is processed by Qwen3.5-27B to remove the post-conclusion continuation; (3) Heuristic, indicating the heuristic method proposed by \cite{li2026role} and (4) HCC, indicating the proposed method in this paper. 
We evaluate the resulting SFT models on three reasoning benchmarks: MATH500, AMC23, and GSM8K.
The primary evaluation metric is pass@1.

\subsection{Main Results}
\noindent \textbf{Overall comparisons.}
Table~\ref{tab:main_results} reports the main SFT results using different processed versions of the same answer-correct long-CoT traces. 
From the results, we can draw several conclusions: 
(1) Across different training traces and models, answer-preserving removal of the editor-identified post-conclusion continuation improves downstream SFT outcomes over training on the original traces.
This provides interventional evidence that the post-conclusion continuation is training-unfavorable in these settings.
(2) HCC achieves performance close to the 27B editor-processed reference, and in several cases even surpasses it.
For example, in the LLaMA3.2-3B setting, HCC obtains an average score of $45.2$ on $\{T\}_{Q}$ and $49.3$ on $\{T\}_{R}$, closely matching the editor-processed results of $45.7$ and $49.8$.
HCC also outperforms the editor reference on MATH500 under $\{T\}_{Q}$ and on GSM8K under $\{T\}_{R}$.
These results indicate that a lightweight boundary proxy can recover much of the benefit of large-model delete-only trace editing.
(3) Compared with heuristic truncation, HCC-processed traces are both more accurate and substantially shorter in response length.
The comparison with heuristic truncation suggests that length reduction alone may not fully explain the gains, although length-controlled interventions would be needed for a complete separation.

\subsection{More Experiments}

\begin{figure}[htb!]
  \centering
  \includegraphics[width=0.48\textwidth]{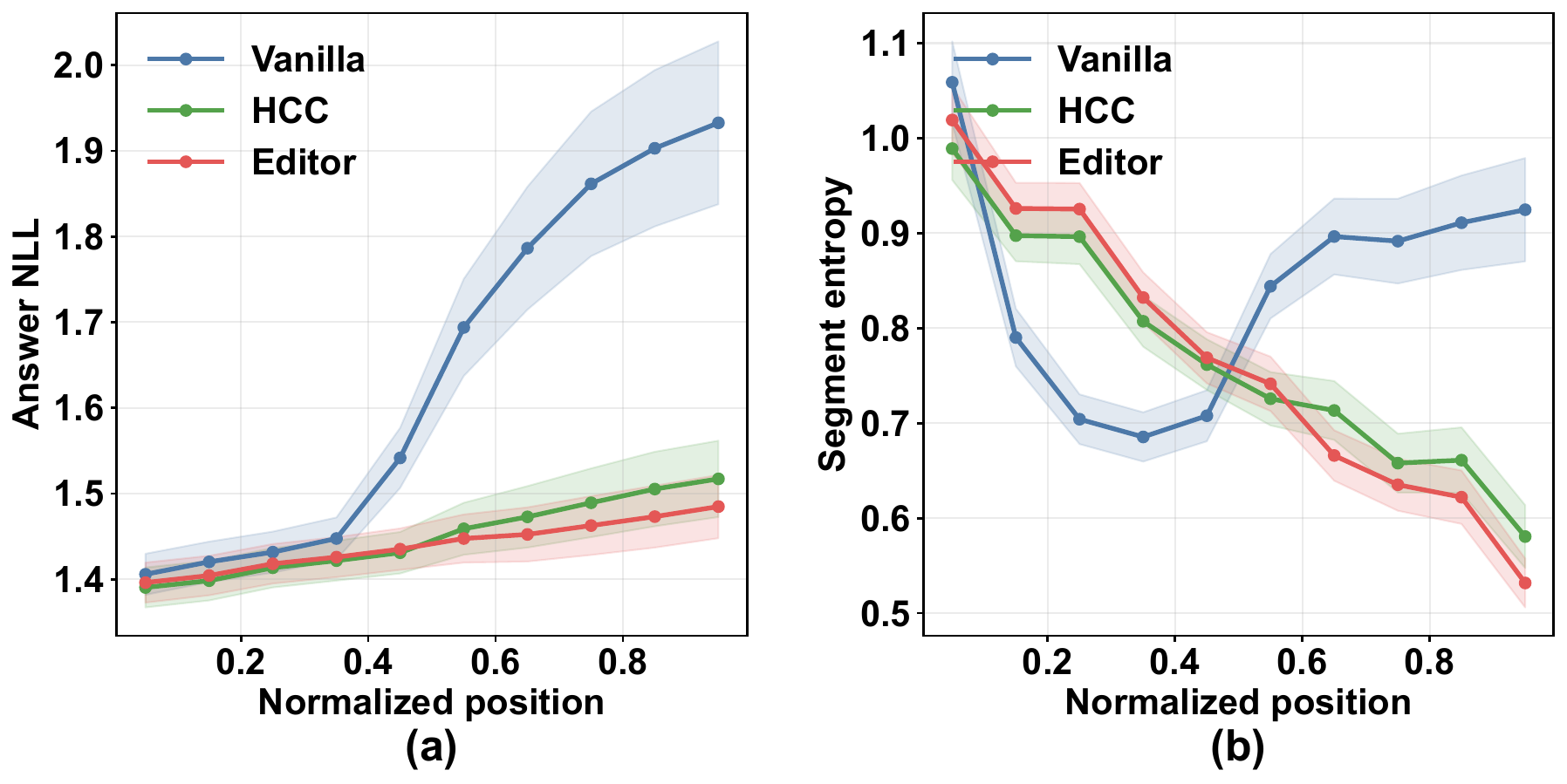}
  \vspace{-4mm}
  \caption{
  Post-training uncertainty diagnostics for generated reasoning traces.
  }
  \label{fig:uncertainty-dynamics}
\end{figure}

\paragraph{Analysis of uncertainty dynamics.}
Figure~\ref{fig:uncertainty-dynamics} compares the reasoning dynamics of models trained on original, HCC-processed, and editor-processed traces.
In Figure~\ref{fig:uncertainty-dynamics} (a), Vanilla shows a sharp late-stage increase in answer NLL, suggesting that generated reasoning becomes less favorable under evaluator-based answer recoverability.
In contrast, both HCC and Editor keep answer NLL much more stable across the reasoning process, suggesting that post-conclusion continuation removal improves answer-support consistency under this diagnostic.
Figure~\ref{fig:uncertainty-dynamics} (b) further shows the segment-level uncertainty pattern.
Vanilla first reduces entropy but then exhibits a clear entropy rebound in the later reasoning bins, which is consistent with re-entering an unstable post-conclusion continuation.
By comparison, HCC and Editor continue to reduce segment entropy in the later stage and produce highly similar curves.
These curves provide post-training diagnostic evidence consistent with Section~\ref{sec:illustration} pattern.
The similarity between HCC and Editor curves suggests that the lightweight proxy approximates the diagnostic behavior of delete-only edited traces.

\begin{figure}[htb!]
  \centering
  \includegraphics[width=0.48\textwidth]{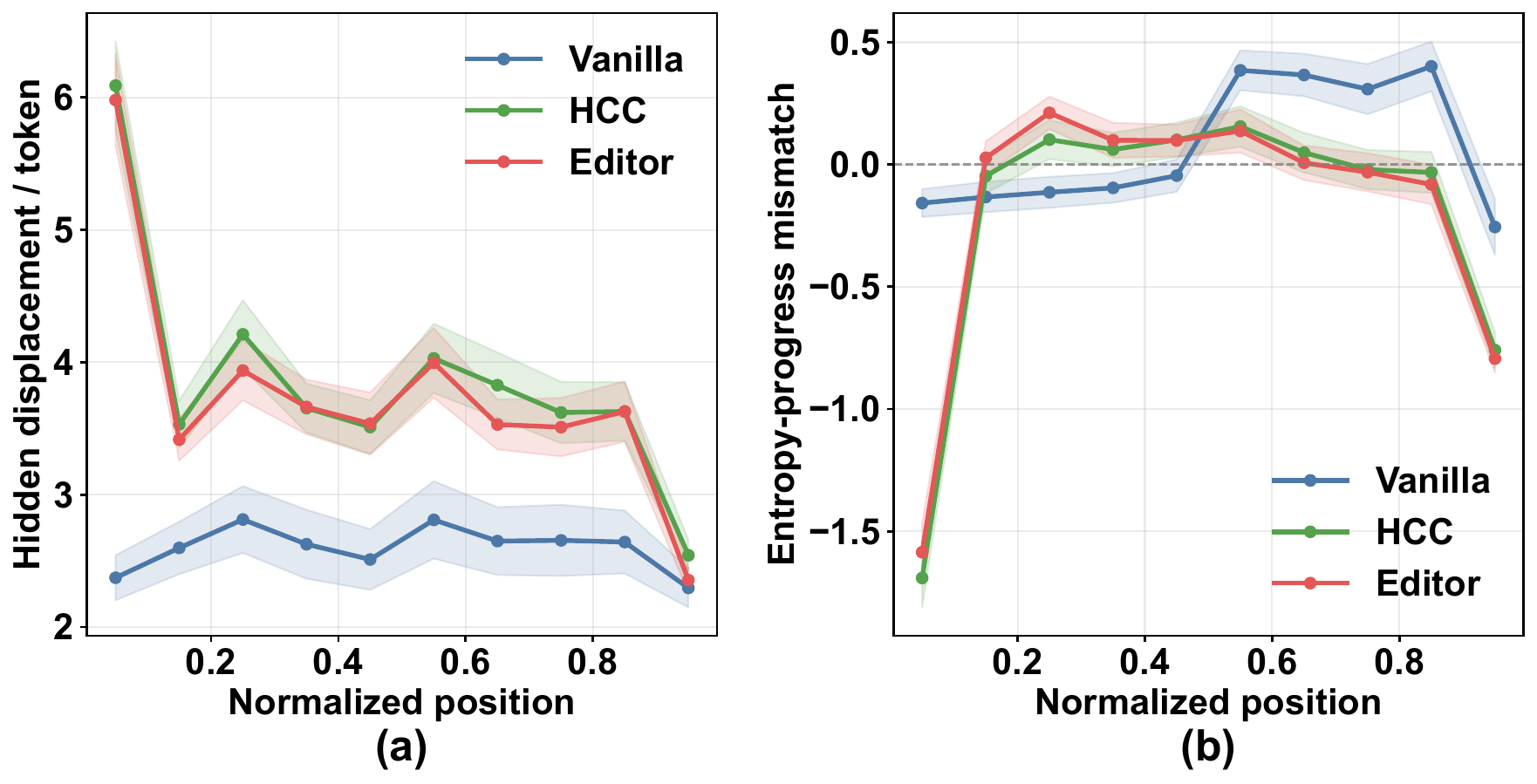}
  \vspace{-4mm}
  \caption{
  Operational uncertainty-progress diagnostics in generated reasoning traces.
  }
  \vspace{-4mm}
  \label{fig:geometry-mismatch}
\end{figure}

\paragraph{Analysis of geometric dynamics.}
Figure~\ref{fig:geometry-mismatch} examines whether processed traces lead to stronger hidden-state progress under the chosen proxy.
In Figure~\ref{fig:geometry-mismatch} (a), HCC and Editor produce larger token-normalized hidden displacement than Vanilla, suggesting stronger state movement per generated token.
Figure~\ref{fig:geometry-mismatch} (b) shows that Vanilla exhibits a clearer positive entropy-progress mismatch in the middle and late stages, where uncertainty is not matched by sufficient geometric progress.
By contrast, HCC and Editor keep this mismatch closer to zero and reduce it near the end.
These curves provide post-training diagnostic evidence consistent with Section~\ref{sec:illustration} pattern, and again suggest that HCC approximates the delete-only editor behavior.

\paragraph{Analysis of reinforcement learning.}
We further examine whether HCC-processed SFT provides a stronger initialization for subsequent GRPO training.
Using LLaMA3.2-3B-Instruct trained on $\{T\}_{Q}$, we apply GRPO to the checkpoints obtained from Vanilla SFT and HCC-based SFT, and evaluate performance across different RL steps.
As shown in Table~\ref{tab:step_comparison}, the model initialized from HCC-based SFT consistently outperforms the Vanilla counterpart at each evaluated step.
The model initialized from HCC-based SFT maintains higher performance across the evaluated RL steps in this setting.
This suggests that harmful continuation removal can provide a stronger SFT initialization.
More broadly, the effect of SFT data processing can persist into subsequent RL.

\begin{table}[htb!]
  \centering
  \small
  \setlength{\tabcolsep}{3pt}
  \renewcommand{\arraystretch}{1.05}
  \caption{Comparison across reinforcement steps.}
\begin{tabular}{ccccccc}
\toprule
\textbf{Method} & \textbf{Dataset} &
\textbf{0} & \textbf{10} & \textbf{20} & \textbf{30} & \textbf{40} \\
\midrule
\multirow{2}{*}{Vanilla}
& MATH500 & 29.8 & 34.8 & 35.2 & 36.6 & 36.4 \\
& GSM8K   & 69.0 & 71.6 & 72.7 & 73.1 & 73.3 \\
\midrule
\multirow{2}{*}{HCC}
& MATH500 & 43.2 & 45.8 & 46.0 & 46.2 & 49.4 \\
& GSM8K   & 75.1 & 75.4 & 75.7 & 77.7 & 77.0 \\
\bottomrule
\end{tabular}
\label{tab:step_comparison}
\end{table}


\begin{table}[htb!]
  \caption{
Comparison of random cut and HCC-based method on LLaMA3.2-3B-Instruct.
}
\centering
\small
\setlength{\tabcolsep}{4pt}
\begin{tabular}{ccccc}
\toprule
\textbf{Method} & \textbf{MATH500} & \textbf{AMC23} & \textbf{GSM8K} &\textbf{Avg.} \\
\midrule
Random Cut & 31.4 & 7.5 & 48.1 & 29.0 \\
HCC & 47.6 & 22.5 & 77.8 & 49.3 \\
\bottomrule
\end{tabular}
\label{tab:random-cut}
\end{table}

\paragraph{Analysis of Random Cut.}
We further introduce a random cut baseline to rule out the possibility that the improvement mainly comes from shorter responses.
To align it with HCC, random cut preserves the final answer, removes a sentence-complete suffix from the reasoning trace, and controls the removed length to match the average truncation length of HCC.
As shown in Table~\ref{tab:random-cut}, random cut is consistently inferior to HCC on MATH500, AMC23, and GSM8K, yielding an average score of only 29.0 compared with 49.3 for HCC.
This large gap suggests that arbitrary length reduction is not a reliable solution.
Since random cut does not identify whether the reasoning has already concluded, it may discard necessary intermediate steps and damage the reasoning chain.
HCC instead removes post-conclusion continuation, thereby reducing redundant tail reasoning while preserving the core answer-supporting process.

\begin{figure}[htb!]
  \centering
  \includegraphics[width=0.48\textwidth]{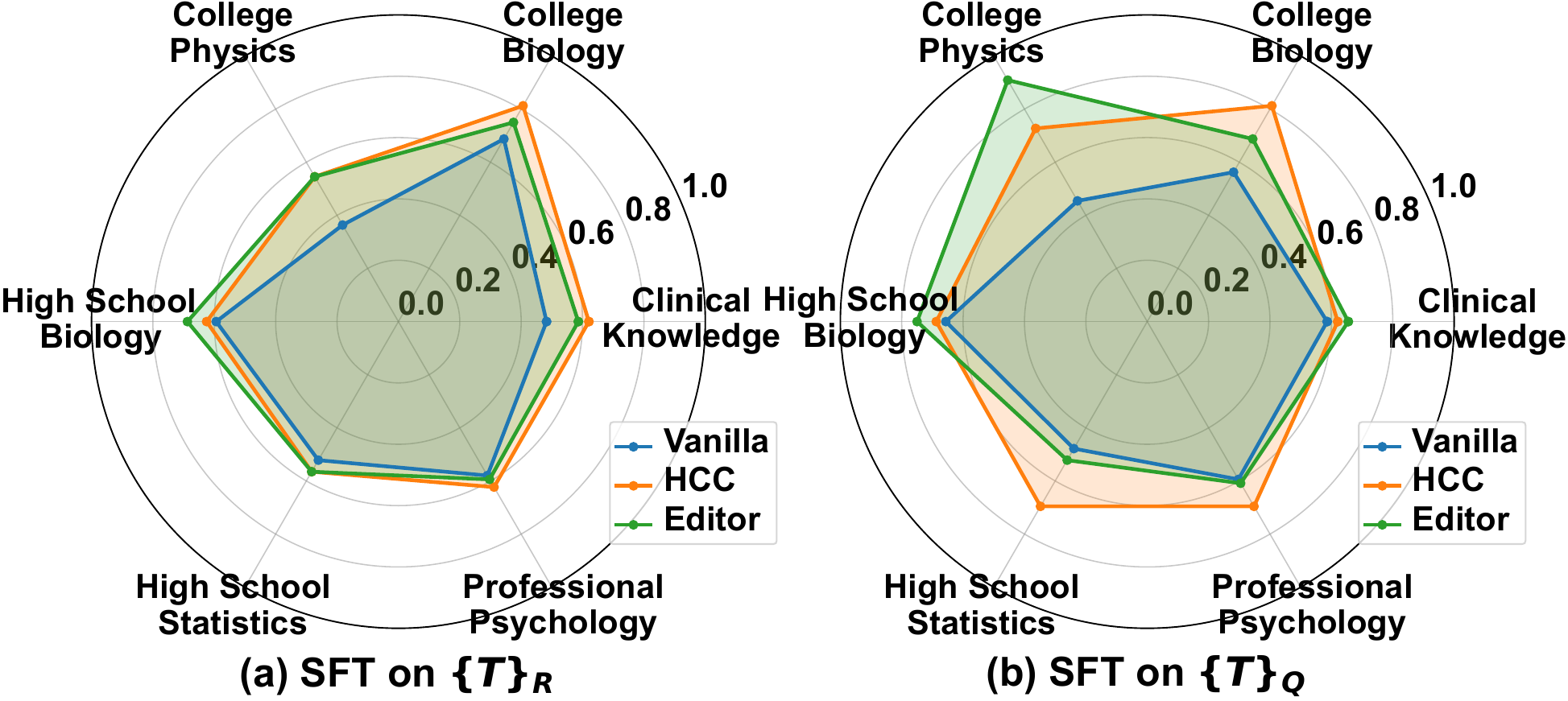}
  \caption{
  Visualization of the performance of LLaMA3.2-3B-Instruct on selected MMLU subjects.
  }
  \label{fig:mmlu-radar}
\end{figure}

\paragraph{Analysis of MMLU datasets.}
We further examine whether the SFT improvements transfer to several non-mathematical evaluation subjects on MMLU \cite{hendryckstest2021}. 
As shown in Figure~\ref{fig:mmlu-radar}, we select 6 subjects that require different types of knowledge, including college physics, college biology, clinical knowledge, professional psychology, high school statistics, and high school biology.
From the results, we can see that HCC-based SFT outperforms Vanilla-based SFT across all subjects, and achieves comparable performance to the editor-based reference. 
These results suggest that models trained on HCC-processed mathematical traces can retain or improve performance on selected out-of-domain knowledge-intensive evaluations.
They do not by themselves establish that the same harmful continuation pattern appears in non-mathematical training traces.
The overall performance of different models and different SFT data settings are shown in the Appendix.

\section{Conclusion}
We studied post-conclusion continuation in answer-correct long-CoT SFT traces, where generation continues after the answer appears sufficiently supported.
Using a delete-only editor, we constructed answer-preserving post-conclusion continuation removal and observed improved downstream SFT outcomes, suggesting that the removed continuation is training-unfavorable in our setting.
We therefore refer to this empirically supported phenomenon as harmful continuation.
We further characterized the removed continuation through uncertainty and operational hidden-state analyses, revealing an uncertainty--geometry mismatch.
Based on this, we instantiated HCC as a lightweight boundary proxy for approximating editor-identified post-conclusion continuation removal.

\section*{Limitations}
The delete-only editor provides an operational intervention, not a ground-truth oracle of harmfulness; its labels should be understood as editor-identified post-conclusion continuation boundaries.
Our measurements are diagnostic proxies rather than causal proof of training harm.
Finally, HCC approximates the editor-identified removal boundary rather than intrinsic harmfulness, and finer component-level attribution remains future work.

\bibliography{custom}

\clearpage
\appendix

\section{Implementation Details}
We implement SFT with LLaMAFactory.
All models are fine-tuned with the AdamW optimizer using a learning rate of $1\times10^{-5}$.
For fair comparison, Vanilla, Heuristic, Editor, and HCC use the same training configuration, and only differ in the SFT traces used for supervision.
During evaluation, we fix the decoding temperature to $0$ to ensure deterministic generation and improve reproducibility.
All benchmark results are reported under this fixed evaluation setting.

\section{Information Bottleneck Motivation for Answer-Sufficiency}
\label{app:ib_proof}

In this appendix, we provide an information-bottleneck motivation for the answer-sufficiency component of HCC.
The goal is to formalize when a reasoning prefix can already support the final answer, and to give an idealized sufficient condition under which a suffix does not change the answer decision under an answer-prediction head.

\subsection{Sequential Conditional Information Bottleneck}
\label{app:seq_ib}

Given a question $Q$, a sentence-level reasoning trace $R=(r_1,\dots,r_T)$, and the final answer $A$, let $R_{\le t}$ and $R_{>t}$ denote the prefix and suffix at step $t$.
For each prefix $R_{\le t}$, we introduce a stochastic bottleneck representation $Z_t$ induced from the contextualized reasoning state $\tilde{h_t}$:
{\small
\begin{equation*}
q_\phi(z_t \mid \tilde{h_t})
=
\mathcal{N}(\mu_t,\Sigma_t).
\end{equation*}
}

The ideal conditional information bottleneck objective aims to preserve answer-relevant information while removing unnecessary dependence on the reasoning prefix:
{\small
\begin{equation*}
\begin{aligned}
\min \quad & I(Z_t;R_{\le t}\mid Q) \\
\mathrm{s.t.} \quad & I(Z_t;A\mid Q) \ \text{is sufficiently large}.
\end{aligned}
\end{equation*}
}
Equivalently, one may optimize the following Lagrangian form:
{\small
\begin{equation*}
\mathcal{J}_{\mathrm{IB}} 
=
- I(Z_t;A\mid Q)
+
\beta I(Z_t;R_{\le t}\mid Q).
\end{equation*}
}

Since exact mutual information is intractable for free-form reasoning traces, we use a variational approximation.
The answer-relevance term is optimized through a variational answer predictor $p_\psi(a\mid z_t,q)$, leading to the negative log-likelihood term:
{\small
\begin{equation*}
-\log p_\psi(a^*\mid z_t,q).
\end{equation*}
}
For the compression term, we introduce a sequential prior $p_\eta(z_t\mid z_{t-1},q)$ and use the following KL penalty as a variational surrogate for the incremental information injected into the bottleneck:
{\small
\begin{equation*}
D_{\mathrm{KL}}
\big(
q_\phi(z_t\mid \tilde{h_t})
\,\|\,
p_\eta(z_t\mid z_{t-1},q)
\big).
\end{equation*}
}
Thus, the practical sequential IB loss is:
{\small
\begin{equation*}
\begin{aligned}
\mathcal{L}_{\mathrm{IB}}
=
\sum_{t=1}^{T}
\Big[
&-\log p_\psi(a^* \mid z_t, q) \\
&+
\beta
D_{\mathrm{KL}}
\big(
q_\phi(z_t \mid \tilde h_t)
\,\|\,
p_\eta(z_t \mid z_{t-1}, q)
\big)
\Big].
\end{aligned}
\end{equation*}
}
This objective encourages $Z_t$ to retain information predictive of the final answer while discouraging unnecessary state complexity.

\subsection{Ideal Answer-Sufficiency Boundary}
\label{app:ideal_sufficiency}

We next formalize an ideal boundary after which the remaining reasoning suffix provides no additional answer information under the bottleneck representation.

\paragraph{Definition.}
The ideal answer-sufficiency boundary is defined as:
{\small
\begin{equation*}
\tau^*
=
\min
\left\{
t:
I(A;R_{>t}\mid Q,Z_t)=0
\right\}.
\end{equation*}
}
This condition means that, once $Q$ and $Z_t$ are given, the suffix $R_{>t}$ provides no additional information about the final answer $A$.
If the set is empty, the trace has no boundary satisfying this ideal condition.

\paragraph{Proposition.}
If
{\small
\begin{equation*}
I(A;R_{>t}\mid Q,Z_t)=0,
\end{equation*}
}
then conditioning on $R_{>t}$ does not change the Bayes-optimal answer distribution given $(Q,Z_t)$.

\paragraph{Proof.}
By the definition of conditional mutual information, the condition
{\small
\begin{equation*}
I(A;R_{>t}\mid Q,Z_t)=0
\end{equation*}
}
implies the conditional independence relation:
{\small
\begin{equation*}
A \perp R_{>t}\mid (Q,Z_t).
\end{equation*}
}
Therefore, for any answer candidate $a$, we have:
{\small
\begin{equation*}
p(a\mid Q,Z_t,R_{>t})
=
p(a\mid Q,Z_t).
\end{equation*}
}
As a result, the Bayes-optimal answer predictor satisfies:
{\small
\begin{equation*}
\begin{aligned}
\arg\max_a p(a\mid Q,Z_t,R_{>t})
&=
\arg\max_a p(a\mid Q,Z_t).
\end{aligned}
\end{equation*}
}
Thus, once $Z_t$ is answer-sufficient in the conditional-independence sense, the suffix $R_{>t}$ is irrelevant to the answer decision under the bottleneck representation.
\qed

\paragraph{Remark.}
This proposition establishes an answer-sufficiency condition only.
It does not imply that deletion necessarily improves SFT.
The empirical benefit studied in the main paper additionally depends on whether the answer-sufficient suffix introduces high local uncertainty or weak geometric progress as supervision.

\section{Experimental Settings}
\label{app:exp_settings}
\subsection{Uncertainty Metrics}
\label{app:uncertainty_metrics}

We compute uncertainty metrics with a fixed evaluator model.
Each answer-correct trace is split into sentence-level units, and the final boxed answer is used as the answer target.
When scoring intermediate reasoning text, standalone boxed-answer strings are removed to reduce trivial answer leakage.

For sentence-level uncertainty, each sentence $r_t=(y_1,\ldots,y_m)$ is scored under its preceding context $P_{t-1}$.
We report token-averaged NLL and predictive entropy:
{\small
\begin{equation*}
\mathrm{NLL}_{\mathrm{sent}}(r_t)
=
-\frac{1}{m}\sum_{i=1}^{m}
\log p(y_i \mid P_{t-1},y_{<i}),
\end{equation*}
}
{\small
\begin{equation*}
\mathrm{Ent}_{\mathrm{sent}}(r_t)
=
\frac{1}{m}\sum_{i=1}^{m}
H\!\left(p(\cdot \mid P_{t-1},y_{<i})\right).
\end{equation*}
}

For answer-level uncertainty, we measure how a reasoning prefix affects recovery of the boxed answer $a^*=(a_1,\ldots,a_L)$.
Given the prefix $P_t$ after appending sentence $r_t$, we compute:
{\small
\begin{equation*}
\mathrm{NLL}_{\mathrm{ans}}(P_t)
=
-\frac{1}{L}\sum_{i=1}^{L}
\log p(a_i \mid P_t,a_{<i}).
\end{equation*}
}
Answer entropy is computed analogously over the answer-token positions.
We define answer-NLL reduction as:
{\small
\begin{equation*}
\Delta_{\mathrm{ans}}(t)
=
\mathrm{NLL}_{\mathrm{ans}}(P_{t-1})
-
\mathrm{NLL}_{\mathrm{ans}}(P_t).
\end{equation*}
}
A larger value means that appending $r_t$ makes the final answer easier to recover under the evaluator.

For segment-wise plots, sentences are appended along the original complete trace, while the x-axis is normalized separately within retained reasoning and editor-removed continuation.
For boundary-level plots, $K_1$ and $K_T$ denote the first and last retained sentences, and $C_1$ and $C_T$ denote the first and last editor-removed sentences.

\subsection{Geometric Metrics}
\label{app:geometric_metrics}

We compute geometric metrics from hidden states at sentence boundaries.
Let $h_t$ denote the evaluator hidden state after consuming the prefix ending at sentence $r_t$.
The local state update is:
{\small
\begin{equation*}
\Delta h_t = h_t - h_{t-1}.
\end{equation*}
}

Hidden displacement measures the size of this update:
{\small
\begin{equation*}
D_t = \|\Delta h_t\|_2.
\end{equation*}
}
Forward progress measures the projection of the local update onto the remaining direction toward the terminal state of the analyzed trace:
{\small
\begin{equation*}
G_t
=
\frac{
\langle \Delta h_t, h_T-h_{t-1} \rangle
}{
\|h_T-h_{t-1}\|_2+\epsilon
}.
\end{equation*}
}
This is an operational proxy for terminal-directional hidden-state progress, not a direct measurement of the true reasoning process.

Progress efficiency is defined as:
{\small
\begin{equation*}
E_t
=
\frac{G_t}{D_t+\epsilon}.
\end{equation*}
}
To control for sentence length, we also report token-normalized variants:
{\small
\begin{equation*}
D_t^{\mathrm{tok}}=\frac{D_t}{n_t},
\qquad
G_t^{\mathrm{tok}}=\frac{G_t}{n_t},
\end{equation*}
}
where $n_t$ is the token length of $r_t$.

Curvature is used only as an auxiliary direction-change diagnostic:
{\small
\begin{equation*}
\mathrm{Curv}_t
=
1-
\frac{
\langle \Delta h_{t-1},\Delta h_t\rangle
}{
\|\Delta h_{t-1}\|_2\|\Delta h_t\|_2+\epsilon
},
\qquad t>1.
\end{equation*}
}

For paired comparisons, we average each metric within retained reasoning and editor-removed continuation for each example.
The paired difference is:
{\small
\begin{equation*}
\Delta
=
\mathrm{Mean}_{\mathrm{removed}}
-
\mathrm{Mean}_{\mathrm{retained}}.
\end{equation*}
}
We report group means, the fraction of examples where the removed continuation is lower or higher, and the 95\% confidence interval of $\Delta$.


\section{Additional Experiments}
\label{app:additional_experiments}

\subsection{Additional Analysis of Harmful Continuation Diagnosis}
\label{app:additional_diagnosis}

\begin{figure}[htb!]
  \centering
  \includegraphics[width=0.48\textwidth]{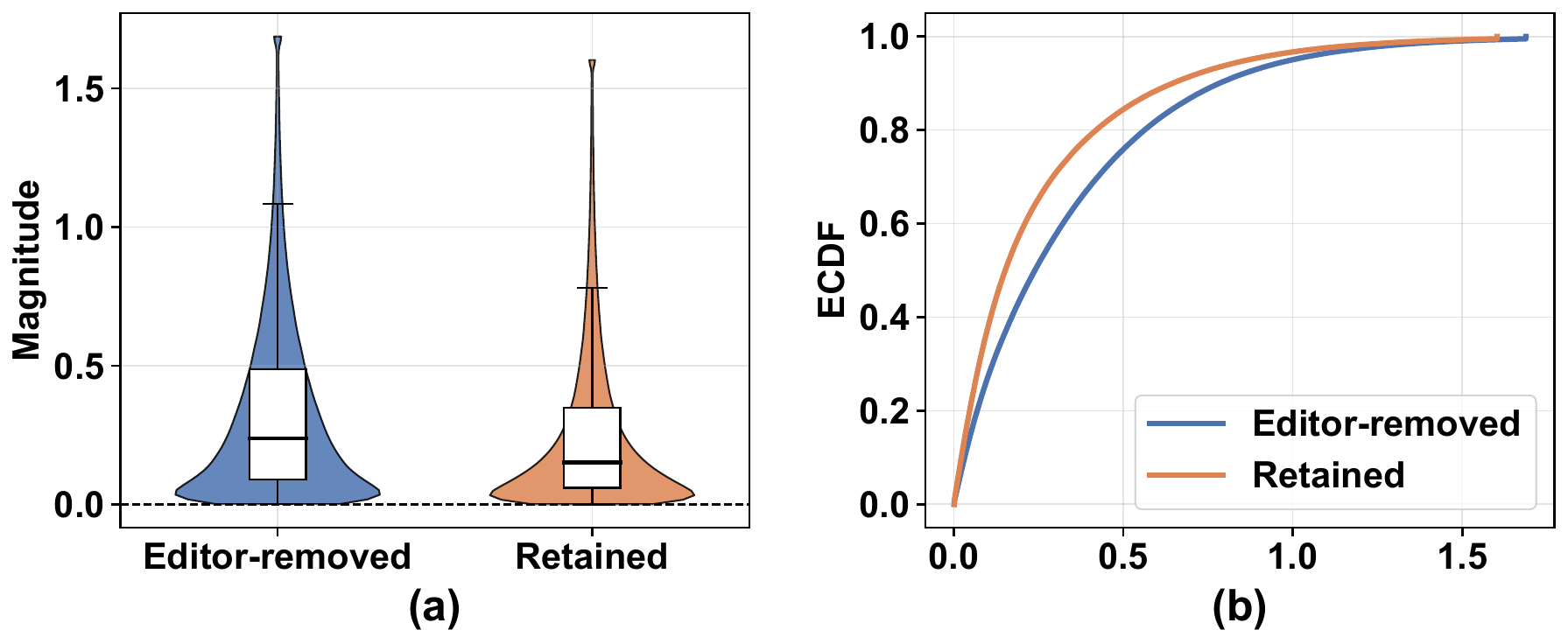}
  \caption{
  Additional uncertainty-side diagnosis.
  (a) NLL perturbation induced by retained reasoning and editor-removed continuation.
  (b) ECDF of average log-probability perturbation.
  Editor-removed continuation induces larger answer-level perturbations.
  }
  \label{fig:app-answer-perturbation}
\end{figure}

\paragraph{Uncertainty View.}
Figure~\ref{fig:app-answer-perturbation} compares the answer-level perturbation induced by retained reasoning and editor-removed continuation.
Editor-removed continuation shows larger NLL perturbation and a right-shifted distribution of average log-probability perturbation.
This suggests that the removed harmful continuation is not simply irrelevant text, but remains answer-conditioned and introduces stronger instability to evaluator-based final-answer prediction.

\begin{figure}[htb!]
  \centering
  \includegraphics[width=0.48\textwidth]{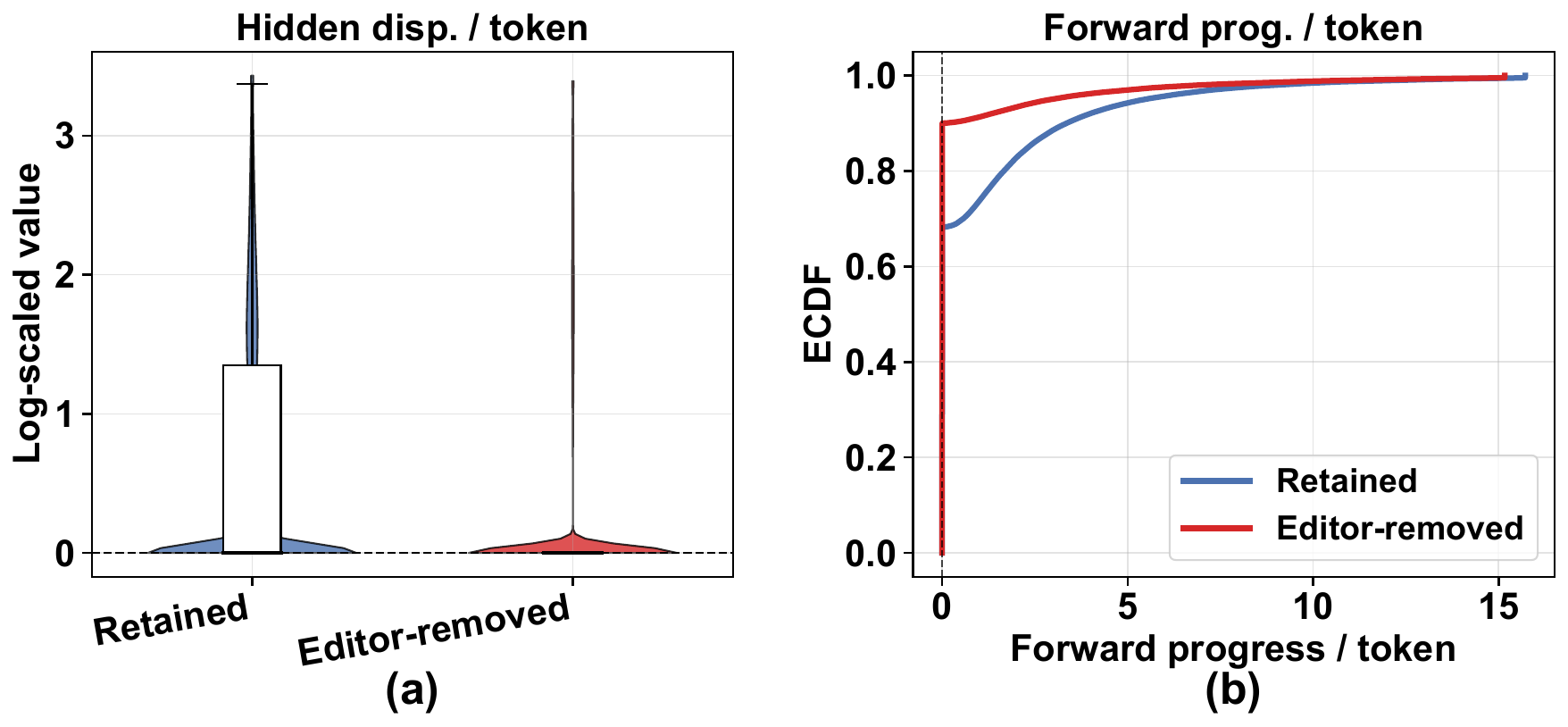}
  \caption{
  Additional geometry-side diagnosis.
  (a) Token-normalized hidden displacement of retained reasoning and editor-removed continuation.
  (b) ECDF of token-normalized forward progress.
  Editor-removed continuation is more concentrated in low-progress regions under the operational proxy.
  }
  \label{fig:app-geometry-diagnosis}
\end{figure}

\paragraph{Geometric View.}
Figure~\ref{fig:app-geometry-diagnosis} further compares the geometric behavior of the two groups.
Retained reasoning induces larger token-normalized hidden displacement, while editor-removed continuation is more concentrated in the low forward-progress region.
This is consistent with the view that editor-removed continuation corresponds to a low-progress phase: it can still affect answer prediction, but does not provide comparable representation-level state movement toward the terminal reasoning state.
Together, these additional results support the uncertainty--geometry mismatch diagnosis of harmful continuation.

\begin{table*}[t]
  \centering
  \small
  \setlength{\tabcolsep}{8pt}
  \renewcommand{\arraystretch}{1.08}
  \caption{Comparison of different methods across backbone models on the MMLU dataset. 
  $\{T\}_{Q}$ and $\{T\}_{R}$ denote SFT trajectories from Qwen-style and R1-style long-CoT sources, respectively.}
  \label{tab:backbone_comparison}
  \begin{tabular}{lcccc}
    \toprule
    \multirow{2}{*}{\textbf{Method}}
    & \multicolumn{2}{c}{\textbf{LLaMA3.2-3B-Instruct}}
    & \multicolumn{2}{c}{\textbf{Qwen3-30B-A3B-Thinking-2507}} \\
    \cmidrule(lr){2-3} \cmidrule(lr){4-5}
    & $\{T\}_{Q}$ & $\{T\}_{R}$
    & $\{T\}_{Q}$ & $\{T\}_{R}$ \\
    \midrule
    Vanilla   
    & 60.1 & 58.0
    & \underline{85.4} & 84.6 \\
    Heuristic 
    & 59.8 & 60.1
    & 85.0 & 85.3 \\
    Editor    
    & 61.1 & 60.6
    & 85.2 & \underline{85.4} \\
    HCC       
    & 62.9 & 60.8
    & \textbf{85.7} & \textbf{85.7} \\
    \bottomrule
  \end{tabular}
  \vspace{-2mm}
\end{table*}

\subsection{Additional analysis of Test Datasets}

\begin{table}[htb!]
\centering
\small
  \setlength{\tabcolsep}{4pt}
  \renewcommand{\arraystretch}{1.05}
\caption{
HCC-based self-consistency diagnostics. Phase, Sent. ratio, and Len. denote the occurrence rate of outputs matching the HCC removable-continuation pattern, the corresponding sentence-level ratio, and the average token length, respectively. These metrics are detector-based consistency measures rather than independent evidence of harmfulness.
}
\begin{tabular}{ccccc}
\toprule
\textbf{Dataset} & \textbf{Method} 
& \textbf{Phase} $\downarrow$
& \textbf{Sent. ratio} $\downarrow$
& \textbf{Len.} \\
\midrule
\multirow{4}{*}{GSM8K}
& Vanilla         & 81.73 & 51.84 & 94.54 \\
& Heuristic       & 93.63 & 59.91 & 96.19 \\
& Editor   & 63.53 & 24.14 & 50.85 \\
& HCC & \textbf{60.42} & \textbf{19.45} & \textbf{48.38} \\
\midrule
\multirow{4}{*}{MATH500}
& Vanilla         & 93.40 & 59.51 & 74.75 \\
& Heuristic       & 93.60 & 60.06 & 73.99 \\
& Editor   & 63.60 & 28.53 & 73.89 \\
& HCC & \textbf{61.80} & \textbf{27.35} & \textbf{66.86} \\
\bottomrule
\end{tabular}
\label{tab:hcc-predicted-bad-phase}
\end{table}

\begin{figure*}[htb!]
  \centering
  \includegraphics[width=0.95\textwidth]{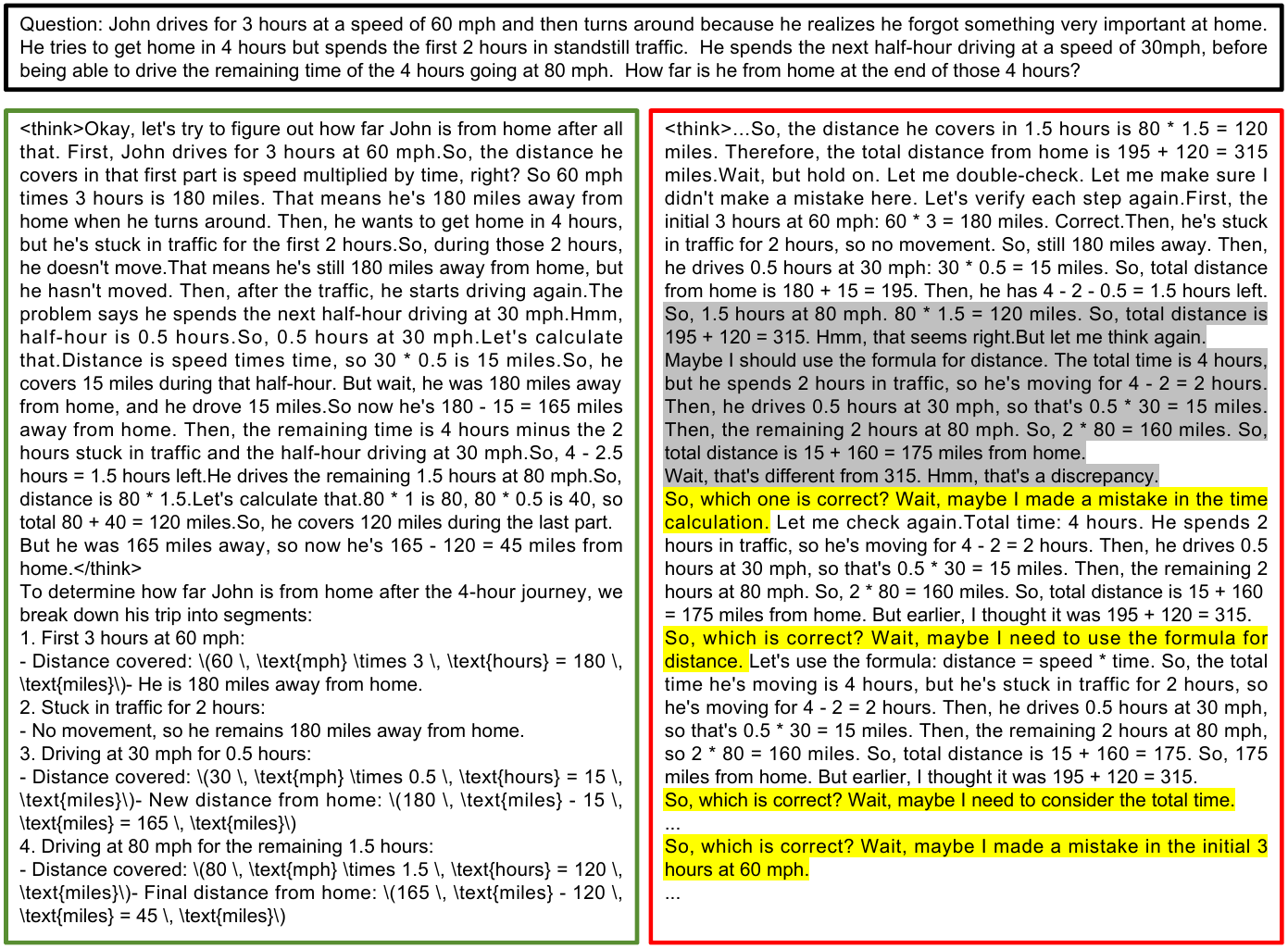}
\caption{
Case study of harmful continuation after SFT.
The left part indicates the reasoning process of a model trained on HCC-processed traces, while the right part shows the reasoning process of a model trained on original traces. We use grey and yellow highlights to indicate the conflicting reasoning and the inefficient reasoning loop, respectively.
  }
  \label{fig:case-study}
\end{figure*}

\paragraph{Case study.}
Figure~\ref{fig:case-study} presents a qualitative example comparing the same base model trained with HCC-processed traces and original long-CoT traces.
The HCC-trained model quickly identifies the correct reasoning path, computes the remaining distance, and outputs the correct solution without entering a long verification loop.
In contrast, the Vanilla-trained model first reaches the correct answer, but then derives a conflicting result from an alternative calculation, as shown in the gray region.
After this conflict appears, the model repeatedly compares the two answers and rechecks different parts of the solution, entering a low-efficiency reasoning loop highlighted in yellow.
The response eventually exhausts the token budget without producing a successful final answer.
This example qualitatively illustrates a behavior consistent with the harmful continuation pattern: the model may continue uncertain, low-progress verification even after a sufficient answer has already been reached.
It suggests that HCC-processed supervision can reduce such continuation patterns.

\paragraph{HCC-based self-consistency diagnostic.}
Table~\ref{tab:hcc-predicted-bad-phase} evaluates whether model outputs after SFT match the removable post-conclusion continuation pattern learned by HCC.
For a fair cross-source test, we use the HCC proxy trained on $\{T\}_{Q}$ to analyze outputs from models trained on $\{T\}_{R}$-based data.
This analysis should be interpreted as a detector-based self-consistency diagnostic rather than an independent measurement of causal harmfulness.
Under this diagnostic, models trained on HCC-processed traces produce fewer outputs that are classified as containing removable post-conclusion continuation.
For example, HCC reduces the sentence-level detected continuation ratio from $51.84\%$ to $19.45\%$ on GSM8K and from $59.51\%$ to $27.35\%$ on MATH500.

\begin{table}[htb!]
\centering
\small
\setlength{\tabcolsep}{4pt}
\begin{tabular}{ccccc}
\toprule
\textbf{Method} & \textbf{\# Params} & \textbf{MACs} & \textbf{FLOPs} & \textbf{Rel. MACs} \\
\midrule
Qwen3.5-27B & 27.0B & 137.1T & 274.3T & $\sim$54.2$\times$ \\
HCC & 498.0M & 2.5T & 5.1T & $1.0\times$ \\
\bottomrule
\end{tabular}
\caption{
Computational cost comparison given similar input lengths of Qwen3.5-27B and HCC.
}
\label{tab:hcc-qwen27b-cost}
\end{table}
\paragraph{Analysis of Computational Costs.}
Table~\ref{tab:hcc-qwen27b-cost} compares the computational cost of the 27B offline editor and our HCC proxy given similar input lengths.
HCC uses only $498$M parameters, which is about $1.8\%$ of Qwen3.5-27B.
It also requires $2.5$T MACs and $5.1$T FLOPs, compared with $137.1$T MACs and $274.3$T FLOPs for Qwen3.5-27B.
This corresponds to a roughly $54.2\times$ reduction in computation.
These results show that HCC provides a much cheaper proxy for editor-identified harmful continuation boundary approximation, making large-scale SFT trace processing more practical.

\end{document}